\documentclass[sn-basic]{sn-jnl}% Math and Physical Sciences Reference Style
\usepackage[ruled,lined]{algorithm2e}
\usepackage{color}
\usepackage{float}

\usepackage{newfloat}
\usepackage{listings}
\newenvironment{listing}[1][H]{%
   \begin{algorithm}%
  }{\end{algorithm}}

\usepackage[scientific-notation=true]{siunitx}

\usepackage{booktabs}

\usepackage{amscd}
\usepackage{cleveref}

\usepackage[utf8]{inputenc}

\definecolor{codegreen}{RGB}{0,100,0}
\definecolor{codegray}{RGB}{128,128,128}
\definecolor{codepurple}{RGB}{76,0,153}
\definecolor{backcolour}{RGB}{1,1,1}

\input{neurolang_listing}

\newcommand{\nop}[1]{}

\newcommand{\D}{\mathcal{D}}

\newcommand{\eqs}{{\,{=}\,}}

\newcommand{\ras}{\,{\ra}\,}

\newcommand{\cups}{\,{\cup}\,}

\newcommand{\cR}{{\cal R}}

\renewcommand{\setminus}{-}

\newcommand{\ra}{\rightarrow}

\newcommand{\dd}[2]{#1_1,\ldots,#1_{#2}}

\newcommand{\vett}[1]{\mathbf{#1}}

\newcommand{\sol}[2]{\mathit{mods}(#1,#2)}
\newcommand{\dom}{\Delta}
\newcommand{\freshdom}{\Delta_N}
\newcommand{\variables}{\mathcal{V}}

\newcommand{\dep}{\Sigma}

\newcommand{\isa}[1]{\mathit{ISA}}

\newcommand{\head}[1]{\mathit{head}(#1)}
\newcommand{\body}[1]{\mathit{body}(#1)}

\newcommand{\answers}[0]{\mathit{ans}}

\newcommand{\atom}[1]{\mathbf{#1}}

\newcounter{cefalo}

\newcounter{cefalocont}

\newtheorem{example}{Example}
\newtheorem{definition}{Definition}

\def\qed{\hfill{\qedboxempty}
  \ifdim\lastskip<\medskipamount \removelastskip\penalty55\medskip\fi}

\def\qedboxempty{\vbox{\hrule\hbox{\vrule\kern3pt
                 \vbox{\kern3pt\kern3pt}\kern3pt\vrule}\hrule}}

\def\qedfull{\hfill{\qedboxfull}
  \ifdim\lastskip<\medskipamount \removelastskip\penalty55\medskip\fi}

\def\qedboxfull{\vrule height 4pt width 4pt depth 0pt}

\jyear{2021}%

\theoremstyle{thmstyleone}%
\begin{document}

\title[ ]{Scalable Query Answering under Uncertainty to Neuroscientific Ontological Knowledge: The NeuroLang Approach}


\author*[1]{\fnm{Gaston E.} \sur{Zanitti}}\email{gaston.zanitti@inria.fr}

\author[2]{\fnm{Yamil} \sur{Soto}}\email{yamil.soto@cs.uns.edu.ar}

\author[1]{\fnm{Valentin} \sur{Iovene}}\email{valentin.iovene@inria.fr}

\author[3]{\fnm{Maria Vanina} \sur{Martinez}}\email{mvmartinez@dc.uba.ar}

\author[3]{\fnm{Ricardo O.} \sur{Rodriguez}}\email{ricardo@dc.uba.ar}

\author[2]{\fnm{Gerardo} \sur{Simari}}\email{gis@cs.uns.edu.ar}

\author[1]{\fnm{Demian} \sur{Wassermann}}\email{demian.wassermann@inria.fr}

\affil[1]{\orgdiv{Parietal Team}, \orgname{INRIA}, \orgaddress{\street{1 Rue Honoré d’Estienne d’Orves}, \city{Palaiseau}, \postcode{91120}, \state{Ile-de-France}, \country{France}}}

\affil[2]{\orgdiv{Dept.\ of Computer Science and Engineering}, \orgname{Universidad Nacional del Sur (UNS) \& Institute for Computer Science and Engineering (UNS--CONICET)}, \orgaddress{\street{San Andres 800}, \city{Bahia Blanca}, \postcode{8000}, \state{Pcia. de Buenos Aires}, \country{Argentina}}}

\affil[3]{\orgdiv{Dept.\ of Computer Science}, \orgname{Universidad de Buenos Aires (UBA) \& Institute for Computer Science Research (UBA--CONICET)}, \orgaddress{\street{Intendente
Güiraldes, Acceso Pabellón 1 2160}, \city{Ciudad Autonoma de Buenos Aires}, \postcode{1428}, \state{CABA}, \country{Argentina}}}


\abstract{Researchers in neuroscience have a growing number of datasets available to study the brain, which is made possible by recent technological advances. Given the extent to which the brain has been studied, there is also available ontological knowledge encoding the current state of the art regarding its different areas, activation patterns, key words associated with studies, etc. Furthermore, there is an inherent uncertainty associated with brain scans arising from the mapping between voxels---3D pixels---and actual points in different individual brains. Unfortunately, there is currently no unifying framework for accessing such collections of rich heterogeneous data under uncertainty, making it necessary for researchers to rely on ad hoc tools. In particular, one major weakness of current tools that attempt to address this kind of task is that only very limited propositional query languages have been developed. In this paper, we present NeuroLang, an ontology language with existential rules, probabilistic uncertainty, and built-in mechanisms to guarantee tractable query answering over very large datasets. After presenting the language and its general query answering architecture, we discuss real-world use cases showing how NeuroLang can be applied to practical scenarios for which current tools are inadequate.}




\maketitle

\section{Introduction}

Recent technological advances in neuroscience have sparked enormous growth in the amount of datasets---containing text, images, and knowledge graphs---available for analysis of the human brain. To take advantage of the full breadth of this heterogeneous, and often noisy data, a unifying framework is needed. This framework should allow researchers to represent their theories, definitions, and perform inferences on them in a structured, formal way. The main hypothesis of this paper is that a probabilistic ontology language based on existential rules (also known as Datalog+/--) carefully extended with negation and aggregation is a useful tool for such tasks.

An example of central neuroscience use cases requiring the combination of the aforementioned datasets are meta-analysis tools. This application constitutes a fertile ground to show how current knowledge representation advancements can combine heterogeneous datasets, pushing forward neuroimaging research. Meta-analysis is a set of techniques used to combine a finite number of published articles, which often disagree, to infer consensus-based findings~\citep{poldrackbrainMapsCognitive2016}. Hence, its main application is aggregating noisy knowledge across articles in the field.  While recent advances in automated meta-analysis techniques are mostly centered in better representing spatial correlations~\citep{samartsidisCoordinatebasedMetaanalysisNeuroimaging2017}, to the best of our knowledge none have formally addressed expressivity limitations of query languages and the feasibility of a more expressive resolution.

Current standard tools for neuroimaging meta-analysis are Neurosynth~\citep{yarkoniLargescaleAutomatedSynthesis2011} and BrainMap~\citep{lairdBrainMapStrategyStandardization2011}, which harness automatically-extracted as well as manually-curated information present across neuroscientific articles. Briefly, these tools interpret each article as an independent sample of \emph{neuroscientific knowledge}, and then develop query systems centered on study subset selection and posterior probabilistic inference on such subsets. For instance, selecting all studies mentioning ``fear'' and inferring the most common areas of the brain reported as active---i.e., deferentially oxygenated---in such studies. In these tools, queries select a subset of a total of around 15k full-text articles reporting involvement of several brain locations each, and a brain tessellation of 300k cubes, or voxels, then infer commonalities across these articles through maximum likelihood estimations combined with spatial information smoothing. Such queries can express questions like ``{\em Where do articles reporting the term `emotion' show activations?}", or ``{\em Which terms associated with cognitive processes are most likely associated with articles reporting activations in the amygdala?}''. Finally, after the inferential tasks, the obtained probabilities are manipulated and aggregated to frame results into the frequentist language neuroscientists commonly use to communicate the significance of their results~\citep{yarkoniLargescaleAutomatedSynthesis2011,samartsidisCoordinatebasedMetaanalysisNeuroimaging2017}. These meta-analyses are performed in under 30 seconds on a regular laptop computer---however, these tools are limited in terms of the expressivity of their associated query languages.

Neurosynth combines text mining, meta-analysis, and machine learning techniques to generate probabilistic mappings relating text-mined terms with activations in the human brain, but the language to infer these relationships is based on propositional logic. This limitation excludes, for instance, the use of existential quantifiers and negation, forbidding queries such as ``{\em What are the terms most probably mentioned in articles reporting activations in the parietal lobe and in no other brain region}'', which we dub \emph{segregation queries}. Another example of this situation is BrainMap, which has a hand-curated dataset of great precision and an ontology for structuring all this knowledge and annotate the articles. Nonetheless, Brainmap's query system is a very limited propositional logic language that only allows to select terms mentioned in the articles and the leaves of the ontology, which again cannot express segregation queries or harness the full information of neuroscience ontologies---such as CogAt~\citep{poldrackCognitiveAtlasKnowledge2011}---that use open knowledge.

Breaching the expressivity limitations of current approaches and handling heterogeneous data requires tackling several issues: handling noisiness in neuroimaging data and conclusions reported across studies calls for a unifying formalism with probabilistic modeling capabilities; being able to leverage ontological information modeled under the open world assumption; finally, performance cannot be ignored since the amount of information needed to model the human brain is considerable. In short, we need to design a logic-based language capable of: 
(i) performing negation and aggregation; 
(ii) performing probabilistic inference; 
(iii) dealing with open knowledge; 
(iv) post-processing inferred probabilities; and 
(v) dealing with neuroimaging databases having, at least, a similar performance to current meta-analytic tools.

\begin{figure}[H]
\centering
\includegraphics[scale=0.1]{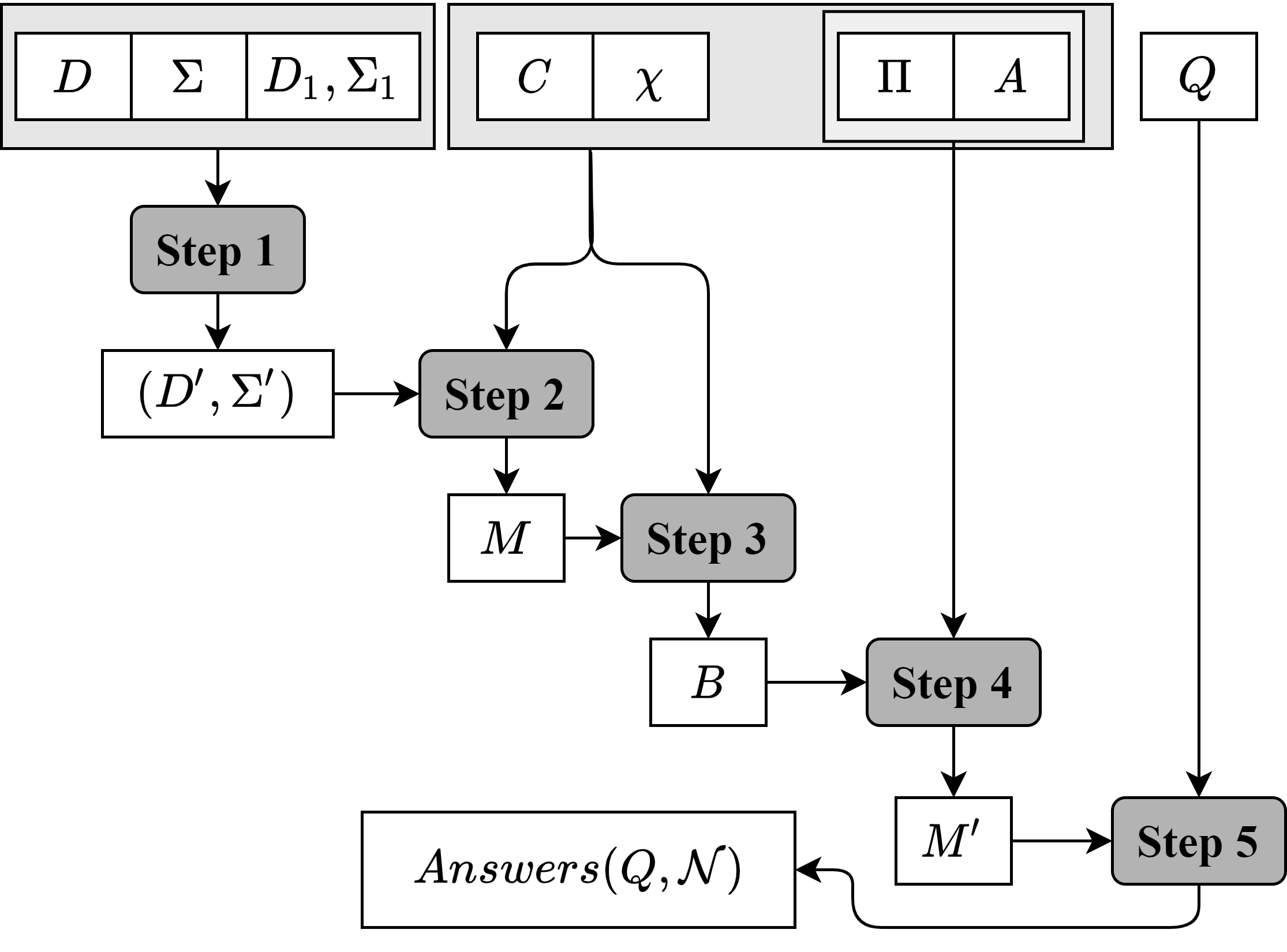}
\caption{Overview of the {\sc NeuroLangQA} algorithm. Step numbers refer to
those described in Algorithm~\ref{alg:NeuroLangQA}}
\label{figure:neurolang}
\end{figure}

Our main proposal in this paper is the development of a subset of Datalog+/--, extended with probabilistic semantics, aggregation, and negation, focused on meta-analytic applications. Such an approach allows us to have a language based on first order logic with negation and existentials (FO$^{\neg\exists}$), enabling more complex queries such as segregation queries or manipulation of open-world information. In all, we produce a language able to express the full breadth of the pipeline needed for meta-analytic applications: from data preprocessing to probabilistic modeling and inference, and finally the post-processing of probabilistic results into images and  reports that are easily interpretable in terms of current reporting used in neuroscience publications. 
Our main contribution is the introduction and evaluation of NeuroLang, a probabilistic language based on Datalog+/-- developed to express and solve rich logic-based queries meeting the functional requirements of neuroimaging meta-analyses.

The rest of this paper is organized as follows: 
Section~\ref{sec:probabilisticmodel} introduces the probabilistic semantics, which is based on a classical possible world approach adopted in many approaches to reasoning under uncertainy;
Section~\ref{sec:neurolang} then formally introduces the NeuroLang language and the {\sc NeuroLangQA} query answering algorithm;
Section~\ref{sec:use-cases} presents a set of real-world use cases showing how our formalism can be applied in neuroscientific research; finally, Section~\ref{sec:Disc-concl}
discusses conclusions.

\section{Basic Probabilistic Ontological Model}
\label{sec:probabilisticmodel}
In this section, we recall the basics on relational databases, conjunctive queries, Datalog, and ontology-mediated query answering (including tuple-generating dependencies and negative constraints), all based on a probabilistic extension with a corresponding query answering semantics. 

We assume an infinite universe of \emph{(data) constants} $\dom$, an infinite set of \textit{(labeled) nulls} $\freshdom$ (used as ``fresh'' Skolem terms)
that are placeholders for unknown values,
and an infinite set of variables~$\variables$. Different constants represent
different values (\emph{i.e., unique name assumption}),
while different nulls may represent the same value.
Sequences of $k\geq 0$ variables, namely $\dd{X}{k}$, are denoted by~$\vett{X}$.

Furthermore, we assume a {\em relational schema} $\cR$, which is a finite set of
{\em predicate symbols}, we also allow built-in predicates (with finite extensions) and equality.
As expected, a~{\em term}~$t$ is a constant, null, or variable.
An {\em atomic formula} (or {\em atom}) $\atom{a}$ has the
form~$p(t_1,\ldots,t_n)$, where~$p$ is an $n$-ary predicate,
and~$t_1,\ldots,t_n$ are terms. We denote with $\mathcal{F}$ the set of all ground atoms built from $\mathcal{R}$ and $\dom$.
A negated atom is of the form $\neg a$ where $a$ is an atom.
We assume that $\mathcal{R} = \mathcal{R}_D \cup \mathcal{R}_P$, with $\mathcal{R}_D \cap \mathcal{R}_P = \emptyset$, containing predicates that refer to deterministic and probabilistic events, respectively. 

A {\em database instance} $D$ for a relational schema~$\mathcal{R}_D$ is a (possibly infinite) set of atoms with predicates from $\mathcal{R}_D$ and arguments from~$\dom$.
On the other hand, let a {\em probabilistic atom} be of the form $\atom{a}:p$, where $p$ is a real number in the interval $[0,1]$ and $\atom{a}$ is an atom with a predicate from $\mathcal{R}_P$. We do not allow negation in probabilistic atoms.

A {\em probabilistic constraint} $c$ has the form
\[
\atom{a}_1:p_1 \,|\, \ldots \,|\, \atom{a}_k:p_k,
\] where $k > 0$, each $\atom{a}_i:p_i$ is a probabilistic atom,  and $\sum p_i \leq 1$. If the $p_i$'s in a probabilistic constraint do not sum to $1$, then there exists also the possibility that none of them happen. The probability of this complementary event is $1 - \sum p_i$.
Given a probabilistic constraint $c = \atom{a}_1:p_1 \,|\, \ldots \,|\, \atom{a}_k:p_k$, we will make use of the notation $\textit{atoms}(c) = \{ \atom{a}_1, \ldots,\atom{a}_k\}$. We will also denote the probability of any atom $\atom{a}$ with $p(\atom{a})$. We have that $p(\atom{a}_i) = p_i$  whenever
$\atom{a}_i:p_i$ belongs to a probabilistic constraint $c$.

%\from{Demian}{All}{Cambiar la propuesta $\sum p_i = 1$ por %$0 < \sum p_i \leq 1$ hace que nuestra semántica incluya %tambien el caso de Tuple-independent databases como como %un conjunto de constraints $c_j$ tal que $\forall j %|\textit{atoms}(c_j)|=1$ y con el mismo símbolo de %predicado. Sin embargo este cambio complica, o al menos %cambia, la definición de ``choices" en Definition 2.}

Given a set of probabilistic constraints $C$, note that each ground atom can only appear in one constraint in $C$.
From a practical point of view, this assumption restricts the number of possible worlds by limiting the potential combinations. \citet[Eq.~5]{vennekensCPlogicLanguageCausal2009} propose more complex semantics where this assumption is relaxed. This approach is similar to {\em probabilistic databases} ~\citep{Susie11} where each tuple comes from a general probability distribution over tuples and inexistence is one of the options. This allows to incorporate
beliefs about the likelihood of tuples and cell values. 
\begin{example}\label{ex:probconstraints}
Consider the following database instance $D$ and a set of probabilistic constraints $C$ (recall that $t_i$ atoms cannot appear in $C$).

\begin{equation}
    \begin{aligned}
    D &=\{t_1(a), t_1(c), t_2(a), t_2(b)\}\\[8pt]
    C &= \left\{ 
    \begin{array}{llcll}
    c_1 = s(a, b)&: 0.3&&&\\
    c_2 = s(b, c)&: 0.7&&&\\
    c_3 = r(b)&:0.4 &\vert & r(c)&:0.1
    \end{array}
\right\}
    \end{aligned}
\end{equation}
\end{example}

\paragraph{Tuple Generating Dependencies}
Given a relational schema~$\mathcal{R}$, a \emph{tuple-gene\-r\-ating dependency (TGD)}
$\sigma$ is a first-order formula of the form:
\[
\forall \vett{X}\forall \vett{Y}\,\Phi(\vett{X}, \vett{Y}) \,{\ra}\,
\exists{\vett{Z}} \,\Psi(\vett{X},\vett{Z}),
\]
where $\Phi(\vett{X},\vett{Y})$
and $\Psi(\vett{X},$ $\vett{Z})$ are conjunctions of atoms over~$\mathcal{R}$
(without nulls), called
the \emph{body} and the \emph{head} of $\sigma$, denoted $\body{\sigma}$ and
$\head{\sigma}$, respectively.
Such~$\sigma$ is satisfied in a database~$D$ for~$\mathcal{R}$ if and only if, whenever there exists a homomorphism $h$ that maps the atoms of
$\Phi(\vett{X},\vett{Y})$ to atoms of $D$, there exists
an extension~$h'$ of
$h$ that maps the atoms of~$\Psi(\vett{X},\vett{Z})$ to atoms of $D$.
All sets of TGDs are finite here and we assume without loss of generality that
every TGD has a single atom in its head.
Furthermore, we say that a TGD $\sigma$ is {\em full} whenever there are no existential 
variables in the head. Let's extend our example further:

\begin{example}\label{ex:tgds}
Based on Example~\ref{ex:probconstraints} we can add the following
set of rules:
\begin{equation*}
\begin{aligned}
\Sigma =\{  & \forall X \; t_1(X) \rightarrow \exists Z \; o(X, Z),\\
            & \forall X \forall Y \; t_2(X) \wedge o(X, Y)\rightarrow t(X),\\
            & \forall X \forall Y \; s(X, Y) \wedge r(Y)\rightarrow w(X, Y)\}\\[6pt]
A       =\{ & \forall X\forall W \; v(X, W) \rightarrow u(X, \max(W))\}\\
\end{aligned}
\end{equation*}
\end{example}

TGDs can be extended to allow negation---in this work we only allow stratified negation~\citep{abiteboulFoundationsDatabases1995} for full TGDs. Furthermore, as shown by the rule in set $A$
in the previous example, we extend the language so aggregation functions can be used in the head of full TGDs~\citep{abiteboulFoundationsDatabases1995}. As we see in the following section, we restrict the syntax of this type of rules so that neither negation nor recursion is allowed.

\begin{definition}
A {\em probabilistic ontology}
$\mathcal{O} \eqs (D, C, \Sigma)$
consists of a database instance $D$, a set $C$ of probabilistic constraints, 
and a set $\Sigma$ of arbitrary TGDs.
\end{definition}

Note that a database instance can be thought of
as a set of probabilistic constraints with only probabilistic atoms, each one annotated with probability~1. Furthermore, the structure $(D, \Sigma)$ corresponds to a 
knowledge base with existential rules as defined in~\cite{CaliGL12}, whenever rules in $\Sigma$ do not involve atoms that appear in probabilistic constraints. 

\paragraph{Semantics}

We take the notion of possible world (or interpretation) of a probabilistic ontology as a subset of $\mathcal{F}$ and we denote with $\Omega$ the set of all possible worlds. Each possible world $\omega \in \Omega$ satisfies the following property:
$$\forall F \in \mathcal{F}: \omega \models F \mbox{ iff } F \in \omega ; \hspace{5mm} \mbox{otherwise } \  \omega \models \neg F$$
This means that $\omega$ is a complete interpretation of every element of  $\mathcal{F}$.
The usual semantics of a classical Datalog program $P$ is the least Herbrand model that contains exactly all ground facts in $P$ plus every ground atom inferred from it, i.e.\ the intersection of all worlds that satisfy $P$. 

However, in the probabilistic case we need to consider a generalization of this semantics so that every ground fact has associated a probability value. According to this idea, we are going to take the models of a set of non-probabilistic ontologies,
induced by total choices, so that they all share the same TGDs but the corresponding database instances differ. 
As mentioned before, in our approach we have two ways of associating probability to facts. In the first one, a fact corresponds to a Boolean random variable that is true with probability $p$ and false with probability $1 - p$. In the second, we interpret facts as multi-valued random variables instead of binary ones. We use probabilistic constraints for representing both, and assume that the facts within the same constraint are mutually exclusive events, where facts in different constraints are mutually independent events. According to this idea, we give the following definition:

\begin{definition}  
Given a probabilistic ontology $\mathcal{O} \eqs (D, C, \Sigma)$, for each $1 \leq j \leq \vert C \vert: c^j = \atom{a}^j_1:p^j_1 \,\vert\, \ldots \,\vert\, \atom{a}^j_k:p^j_k$, with $c^j\in C$,  we have: 
\[
\textit{choices}(c^j) = \{ \atom{a}^j_i \;\vert\; 1\leq i\leq k\} \cup \{\bot_{c^j}\}.
\]
For each $b= \atom{a}^j_i \in c^j$, we have $p(b) = p^j_i$ and  
$p(\bot_{c^j}) = 1 - \sum_{1\leq i \leq k} p^j_i$. The set of total choices for $\mathcal{O}$ is defined as
$\textit{total\_choices}(C) =$
\[
    \{[b_1, \dots ,b_l] \;\vert\; l =\vert C \vert, \\
    1\leq j \leq \vert C \vert: b_j \in\textit{choices}(c^j)\}
\]
The probability of a particular total choice $\lambda \in \textit{total\_choices}(C)$ 
is defined as
$p(\lambda) = \prod^{[b_1,...,b_l] \in \lambda}_{1 \leq j \leq l} p(b_j)$. We use notation $\textit{atoms}(\lambda) = \{ \atom{b}_j \neq \bot_{c^j} \,\vert\, 1\leq j \leq l : [b_1,...,b_l] \in \lambda \}$ and $\textit{atoms}(C) = \bigcup_{\lambda \in\textit{total\_choices}(C)} \textit{atoms}(\lambda)$.
\end{definition}

%Given a probabilistic ontology $\mathcal{O} \eqs (D, C, \Sigma)$, for each constraint $c = \atom{a}_1:p_1 \,|\, \ldots \,|\, \atom{a}_k:p_k$, with $c\in C$,  we have 
%$\textit{choices}(c) = \{ \{\atom{a}_i\} \,|\, 1\leq i\leq k\} \cup \{\bot_c\}$.
%For each $b= \atom{a}_i \in c$, we have $p(b) = p_i$ and  
%$p(\bot_c) = 1 - \sum_{1\leq i \leq k} p_i$.

%A total choice for $\mathcal{O}$, $\textit{total\_choice}(C)$, is defined as
% $\textit{total\_choice}(C) = \bigcup_{c\in C}\{ b \,|\, b \in\textit{choices}(c)\}$.
% The probability of a total choice $\lambda$, denoted
% $p(\lambda) = \prod_{b\in \lambda} p(b)$.

\begin{definition}
Let $\omega$ and $\lambda$ be a possible world and a total choice, respectively. Then, we will say that $\omega$ satisfies $\lambda$, denoted $\omega \models \lambda$, if and only if   $\textit{atoms}(\lambda) \subseteq \omega$. Also, $\| \lambda \|$ will denote the set of possible worlds of a total choice, i.e.\  $\| \lambda \| = \{ \omega \in \Omega \;\vert \; \omega \models \lambda \}$.
\end{definition}

\begin{example}\label{ex:totalchoices}
The set of all total choices for probabilistic ontology
$(D,C,\Sigma)$ from Examples~\ref{ex:probconstraints} and~\ref{ex:tgds} is the following:
\vspace{0.3cm}
\begin{center}
$\begin{array}{lllrll}
    \lambda_1    & = {[}s(a, b),       & s(b,c),      & r(b){]}       & p(\lambda_1)    & = 0.084   \\
    \lambda_2    & = {[}s(a, b),       & \bot_{c_2},  & r(b){]}       & p(\lambda_2)    & = 0.036   \\
    \lambda_3    & = {[}\bot_{c_1},    & s(b, c),     & r(b){]}       & p(\lambda_3)    & = 0.196   \\
    \lambda_4    & = {[}\bot_{c_1},    & \bot_{c_2},  & r(b){]}       & p(\lambda_4)    & = 0.084   \\
    \lambda_5    & = {[}s(a, b),       & s(b,c),      & r(c){]}       & p(\lambda_5)    & = 0.021   \\
    \lambda_6    & = {[}s(a, b),       & \bot_{c_2},  & r(c){]}       & p(\lambda_6)    & = 0.009   \\
    \lambda_7    & = {[}\bot_{c_1},    & s(b, c),     & r(c){]}       & p(\lambda_7)    & = 0.049   \\
    \lambda_8    & = {[}\bot_{c_1},    & \bot_{c_2},   & r(c){]}       & p(\lambda_8)    & = 0.021   \\
    \lambda_9    & = {[}s(a, b),       & s(b,c),      & \bot_{c_3}{]} & p(\lambda_9)    & = 0.105   \\
    \lambda_{10} &  = {[}s(a, b),      & \bot_{c_2},   & \bot_{c_3}{]} & p(\lambda_{10}) & = 0.045   \\
    \lambda_{11} & = {[}\bot_{c_1},    & s(b, c),      & \bot_{c_3}{]} & p(\lambda_{11}) & = 0.245   \\
    \lambda_{12} & = {[}\bot_{c_1},    & \bot_{c_2},   & \bot_{c_3}{]} & p(\lambda_{12}) &  = 0.105   \\
\end{array}$
\end{center}
\end{example}

\smallskip
It is easy to see that $\textit{total\_choices}(C)$ defines a partition on $\Omega$ by using the following equivalence relation on $\Omega \times \Omega$: $\omega \equiv \omega' \ \text{if and only if} \ \forall \lambda \in \textit{total\_choices}(C): \omega \models \lambda \Leftrightarrow \omega' \models \lambda$.

We define the semantics of a probabilistic ontology based on
the semantics of a classical ontology with existential rules (TGDs).
Intuitively, each total choice induces a classical (i.e., non-probabilistic) ontology.

\begin{definition}
Let $\mathcal{O} \eqs (D, C, \Sigma)$, be a probabilistic ontology, and
let $\lambda$ be a total choice of $C$. 
Then, the (non-probabilistic) ontology induced by $\lambda = [b_1, \ldots b_l]$
is defined as $O_\lambda = (D_\lambda, \Sigma)$, with 
$D_\lambda = D \cup \{b_1, \ldots b_l\}$.
\end{definition}

\begin{example}
Based on the total choices from Example~\ref{ex:totalchoices} and
probabilistic ontology $\mathcal{O} = (D,C,\Sigma,)$, each $\lambda_i$ with $1\leq i \leq 12$, induces a non-probabilistic ontology $\mathcal{O_{\lambda_i}} = (D_{\lambda_i}, \Sigma)$
where $D_{\lambda_i} = \D \cup \{b_1,\ldots, b_l\}$ with
$b_k \in \lambda_i$ and $b_k \neq \bot_{c_j}$ for every $c_j \in C$.
\end{example}

We recall the notion of models and satisfaction for classical ontologies in~\cite{CaliGL12}.

\begin{definition}
 Given an ontology $(D,\Sigma)$, the set of \emph{models}, 
denoted $\sol{D}{\dep}$, is the set of all (possibly infinite) databases $B$ such that (i) $D \subset B$, and
(ii) every $\sigma\,{\in}\,\dep$ is satisfied in $B$.
\end{definition}

Note that each $B$ in the above definition can be considered as a possible world under the closed world assumption, i.e.\ every tuple that does not appear in $B$ is false.
It is important to recall that for full TGDs (pure Datalog rules),  
an ontology $(D,\Sigma)$ has a unique least model~\citep{abiteboulFoundationsDatabases1995}.

\begin{definition} 
Let $\mathcal{O}$ be a probabilistic ontology, and $\Phi$ be a conjunction of ground 
atoms built from predicates in $\mathcal{R}$. The probability that $\Phi$  holds in $\mathcal{O}$, denoted
$\textit{Pr}^{\mathcal{O}}(\Phi)$, is the sum of the probabilities of all 
total choices $\lambda$ such that $(D_\lambda, \Sigma) \models \Phi$;
that is, $\textit{Pr}^{\mathcal{O}}(\Phi) = \sum^{\lambda \in \textit{total\_choice}(C)}_{(D_\lambda, \Sigma) \models \Phi}  p(\lambda)$.
\end{definition}

At this point, it is interesting to remark the connection between our approach and the one considered by~\cite{riguzziALLPADApproximateLearning2008,riguzziALLPADApproximateLearning2006}. The Logic Programs with Annotated Disjunctions (LPADs) mentioned in their paper make an implicit treatment of mutually exclusive facts, whereas our approach does it explicitly. In fact, LPADs are more expressive than our language since they use non-Horn clauses. In addition, they use well-founded semantics in order to deal with negation as failure. Both aspects have a computational cost that we wish to avoid.

\paragraph{Semantics for Query Answering}

A {\em conjunctive query (CQ)} over $\mathcal{R}$ has the form $Q(\vett{X})=\exists \Phi(\vett{X},\vett{Y})$,
where 
$\Phi(\vett{X},\vett{Y})$ is a conjunction of atoms 
(possibly equalities, but not inequalities) with the variables $\vett{X}$ and~$\vett{Y}$,
and possibly constants, but without nulls.
%A {\em Bool\-ean CQ (BCQ)} over %$\R$ is a $CQ$ of the form $Q()$,
%often written as the set of all its %atoms, without  quantifiers.
Probabilistic answers to CQs are defined via {\em homomorphisms}, which are
mappings $\mu\colon \dom \cup
\freshdom\cup\variables\rightarrow\dom \cup \freshdom\cup\variables$ such that
(i) $c \in \dom$ implies~$\mu(c)=c$,
(ii)~$c \in \freshdom$ implies~$\mu(c) \in \dom\cup\freshdom$,
and (iii)~$\mu$~is naturally extended to atoms, sets of atoms,
and conjunctions of atoms.

\begin{definition}\label{def:answers}
The set of all {\em probabilistic answers} to a CQ $Q(\vett{X})\,{=}\,\exists\vett{Y}\, \Phi(\vett{X},\vett{Y})$
over a probabilistic ontology $\mathcal{O} \eqs (D, C, \Sigma)$,
denoted with $\answers(Q,D,C,\Sigma)$, or $\answers(Q,\mathcal{O})$,
is a set of pairs
$(t, p_t)$ with $t$ a tuple over $\Delta$ such that there exists
a homomorphism $\mu\colon \vett{X}\cups \vett{Y}\ras \dom \cup \freshdom$
with $\mu(X) = t$ and $(D_\lambda,\Sigma) \models \mu( \Phi(\vett{X},\vett{Y}))$
for all $\lambda \in  \textit{total\_choice}(C)$.
The probability of each tuple $t$ is then $p_t = \sum^{\lambda \in \textit{total\_choice}(C)}_{(D_\lambda, \Sigma) \models \mu( \Phi(\vett{X},\vett{Y}))}  p(\lambda)$.
\end{definition}
%The {\em answer}~to a BCQ $Q()$ (a %CQ without free variables) over 
%$\mathcal{O}$ is~{\em Yes}, %denoted~$\mathcal{O}\,{\models}\, %Q$,
%iff $\answers(Q,D,C,\Sigma) %\,{\neq}\,\emptyset$.

\paragraph{Observations} 

If a probabilistic ontology $\mathcal{O}\eqs (D, C, \Sigma)$ is such that $C$ is empty, then the semantics for (B)CQs as defined above coincides with that for classical ontologies~\citep{CaliGL12}.

Note that query answering under general TGDs for non-probabilistic ontologies is undecidable~\citep{BeVa81}, even when the schema and TGDs are fixed~\citep{CaGK08}. The two problems of CQ and BCQ evaluation under TGDs are \textsc{log\-space}-equi\-valent~\citep{FKMP05,DeNR08}.
As mentioned above,  in the non-probabilistic case, for arbitrary full TGDs there exists exactly one
minimal model~\citep{abiteboulFoundationsDatabases1995} over which $Q$ is evaluated. Furthermore, it has been shown that for full TGDs CQ evaluation can be done in polynomial time in data complexity ({\em i.e., assuming $\sigma$ and $Q$ fixed})~\citep{dantsin2001complexity}.

\section{NeuroLang Programs}
\label{sec:neurolang}

%\begin{figure*}[h!]
%    \centering
%    %\includegraphics[width=\linewidth]{./fi%gures/Arquitectura semántica}
%    \includegraphics[width=.8\textwidth]{./f%igures/NeuroLang Arch for paper}
%    \caption{}\label{figure2}
%\end{figure*}

%If we view the semantics %of the underlying %deterministic
%language as a map from %programs to executions of %the program,
%the semantics of the %probabilistic language %will be a map from
%programs to distributions %over executions.
%\from{Ricardo}{all}{Propongo la siguiente definición alternativa de NeuroLang Program que me parece más clara y consistente con la línea notacional inicial}
%\textcolor{red}{A {\em NeuroLang program} $\KB \eqs (\mathcal{O}, \sigma)$ where $\mathcal{O} = (D, C, \Sigma)$ is probabilistic ontology and $\sigma$ is a mapping $\sigma: \Sigma \mapsto \mathbb{P}$ which associated to each TGD in $\Sigma$ a probabilistic distribution in $\mathbb{P}$, in fact, it want to be a conditional probability}. 
%\from{Ricardo}{all}{La idea de esta definición alternativa es reproducir lo mismo que hacemos con los probabilistic constraints de los átomos. Si son crisp le asignamos 1 y 0 a su negación. Ahora si el TGD es clásico entonces cada instancia toma valor 1. Ahora lo que tenemos es la aplicación de la regla de Bayes para obtener la probabilidad del consecuente a partir de la probabilidad del producto de las probabilidad de los antecedentes y la probabilidad condicional.}

In addition to our model, we assume the existence of a separate schema $\mathcal{T}$, the target schema, that defines the language by means of which users of NeuroLang can query about the probability of certain events.
Predicates in $\mathcal{T}$ have a distinguished term in the $n$-th position (for $n$-ary predicates) reserved exclusively for real numbers in the interval $[0,1]$; i.e., for any predicate $p \in \mathcal{T}$, atoms of the form $p(a_1,\ldots,a_n)$ are such that $a_1, \ldots, a_{n-1}$ are variables or constants from $\Delta$, while $a_n$ is a variable or a constant from $[0,1]$. Below we show an example of how this language is used.

A NeuroLang program $\mathcal{N}$ is comprised of the following components:
\begin{itemize}
\item $D$, $\Sigma$: where $D$ is a set of ground atoms from  $\mathcal{R}_D$, and $\Sigma$ is a set of full TGDs that only use atoms from $\mathcal{R}_D$ and can have recursion and stratified negation.

\item $(D_1, \Sigma_1)$: a classical ontology, where $D_1$ is a set of ground atoms from $\mathcal{R}_D$, $\Sigma_1$ is a set of TGDs that belong to the Sticky fragment~\citep{cali_towards_2012}, and the bodies and heads are atoms built from predicates in $\mathcal{R}_D$.

\item $C$: a set of probabilistic constraints only involving atoms from $\mathcal{R}_P$.

\item $\chi$: a set of full TGDs, whose bodies and heads may contain atoms from $\mathcal{R}_D \cup \mathcal{R}_P$. Neither negation nor recursion is allowed in this set of rules.

\item $\Pi$: a set of {\em probability encoding rules} (PERs) with the following form:
\begin{equation*}
    \sigma^*:  \forall \vett{X} \forall \vett{Y} (\Phi(\vett{X}, \vett{Y})) \rightarrow \psi(\vett{X},    \rho_X)
\end{equation*}
where $\Phi$ is a conjunction of atoms from $\mathcal{R}_D\cup \mathcal{R}_P$,  $\psi$ is an atom in $\mathcal{T}$ and $\rho_X$ is the distinguished term that in this case must be a variable (ranging over the reals in  $[0,1]$).

\item $A$: a set of rules of the form
\begin{equation}
\forall \vett{X} \forall \vett{Y} (\Phi(\vett{X}, \vett{Y}, \vett{Z}) \rightarrow \psi(\vett{X}, \textit{agg}(\vett{Z}))
\end{equation}
where $\Phi$ is a conjunction of atoms in $\mathcal{R}_D \cup \ \mathcal{T}$ and $\textit{agg}$ is an aggregation function (e.g., sum, count, avg, etc.). Neither negation nor recursion is allowed in these rules.
\end{itemize}
Informally, the above sets together provide the following functionalities: 
\begin{enumerate}
\item[(i)] $\Sigma$, $\Sigma_1$, $\mathcal{C}$, and $\chi$ are used by the probabilistic inference mechanism, which applies ontological rules and ultimately associates probabilities to atoms (following the semantics described in Section~2); 

\item[(ii)] $\Pi$ incorporates probabilities as values inside atoms; and 

\item [(iii)] rules in $A$ manipulate these probabilities via aggregation functions to present them as requested by the~user.
\end{enumerate}

\begin{algorithm}[H]
    \SetKwInOut{Input}{Input}
    \SetKwInOut{Output}{Output}
    \Input{NeuroLang program $\mathcal{N} = (D, \Sigma, (C,\chi), (D_1,\Sigma_1), \Pi, A)$ and 
    query $Q(\vett{X}) = \exists \vett{Y} \Phi(\vett{X}, \vett{Y})$}
    \Output{$\answers(Q(\vett{X}),\mathcal{N})$}
    {\bf Step 1:} Obtain database instance $D'$ and set of full TGDs $\Sigma'$ such that $D^{'}=D\cup D_1$ and 
    $\Sigma^{'}$ is the rewriting of $\Sigma$ with respect to $\Sigma_1$. \\
    {\bf Step 2:} \\
    2a: Let $\textit{Aux}$ be the set of TGDs in $\Sigma^{'}$ whose bodies don not depend on 
    $C \cup \chi \cup \Pi$. \\
    2b: Let $M$ the set of ground atoms $a$ such that $(D',\textit{Aux} \cup A) \models a$ \\
    % 2c: Let $C' = \{a: 1 \; | \; a \in M\}$ \\
    {\bf Step 3:} \\
    $B$:= $\emptyset$ \\ 
    \ForEach{PER $\pi \in \Pi$}{
    {\tt \small{// Rule bodies are taken as queries} \\}
    Let $Q_\pi(\vett{X}) = \textit{body}(\pi)$ \quad  \\
    {\tt \small{// Obtain probability values \\// associated with each query answer} \\}
    $\textit{probAnsPairs}$:= \label{line:probAns}
    $\answers(Q_\pi(\vett{X}), (M, C, \chi))$ \quad \\
    \ForEach{$(t,p) \in \textit{probAnsPairs}$}{
    {\tt \small{// Add query answers and PER \\// heads to set $B$} \\}
    Let $h'$ be the instantiation of $\textit{head}(\pi)$ with values from $(t,p)$ \\
    $B$:= $B \cup \{h',Q_\pi(t)\}$ \quad }
    }
    {\bf Step 4:} 
    Let $M'$ the set of ground atoms $a$ such that $(B, (\Sigma' - Aux) \cup A) \models a$ \\
    {\bf Step 5:} 
    Return $\answers(Q(\vett{X}),\mathcal{N})$ computed from atoms in set $M'$.
    \caption{{\sc NeuroLangQA}}
    \label{alg:NeuroLangQA}
\end{algorithm}
\vspace{0.3cm}

Note that PERs are full TGDs that will be used to translate from a source schema to a target one, in the same spirit as source-to-target TGDs for data exchange~\citep{FAGIN200589}.
Effectively, they reify the probability of an atom, given by the semantics, as a term in a new atom that can be further manipulated by other rules. For instance, a set of probabilistic constraints $C=\{s(a, b) : 0.3\}$ will be reified by the PER $\forall X\forall Y s(X, Y) \rightarrow t(X, Y, \rho_X)$ as $\{t(a, b, 0.3)\}$.
On the other hand, for rules in $A$ we incorporate functional symbols $\textit{agg}$ to the distinguished term in $\psi$ to indicate that its value takes the result of applying the function $\textit{agg}$ to all $\rho_X$ that satisfy the body of the rule. Note that users here can define arbitrary rules that manipulate probabilities by means of aggregation functions. %As in the case of PERs, this is also outside of the logic since 
It is defined as a post-processing step that builds a view as defined by the user issuing the query.
We extend notation  $\textit{body}$ and $\textit{head}$ used for TGDs to all types of rules defined in this section. The following is a simple example of query answering using PERs.

\begin{example}\label{ex:neurolangprogram}
Consider the following NeuroLang program $\mathcal{N}$. We add a set of PERs and rules with aggregations.
\begin{equation*}
\begin{aligned}
D_1&=\{t_1(a), t_1(c)\}, \\
\Sigma_1&=\{\forall X t_1(X)\rightarrow \exists Z \; o(X, Z)\},\\[4pt]
D&=\{t_2(a), t_2(b)\}, \\
\Sigma&=\{\forall X \forall Y \ t_2(X) \wedge o(X, Y)\rightarrow t(X)\}, \\[4pt]
C &= \left\{ \begin{array}{llcll}
s(a, b)&: 0.3&&&\\
s(b, c)&: 0.7&&& \\
r(b)&:0.4 & \vert & r(c)&:0.1
\end{array}\right\},\\[4pt]
\chi &= \{\forall X \forall Y\, s(X, Y) \wedge r(Y)\rightarrow w(X, Y)\}\}, \\[4pt]
\Pi &=\{\forall X \forall Y\, w(X, Y)\rightarrow v(X, \rho_X)\}, \\[4pt]
A&=\{\forall X \forall W\, v(X, W)\rightarrow u(\max(W))\},\\[8pt]
Q_1(X, P)& =  v(X,P),t(X),\\
Q_2(X, P)& = v(X,P), u(P).
\end{aligned}
\end{equation*}
Now, the partition of possible worlds used to compute queries $Q_1$ and $Q_2$ is the following (excluding atoms from $D$ and $(D_1, \Sigma_1)$ for clarity, and including probabilities):

\begin{equation*}
\left\{\begin{array}{llllrl}
     \{s(a, b) & s(b,c) & w(a, b) &r(b)&t(a)\} &:0.084   \\
     \{s(a, b) &  & w(a, b) &r(b)&t(a)\} &:0.036   \\
     \{&  s(b, c) & &r(b)&t(a)\} &:0.196   \\
     \{ &  & &r(b)& t(a)\} &:0.084   \\
     \{s(a, b) & s(b,c) & w(b, c) &r(c)&t(a)\} &:0.021   \\
     \{s(a, b) &  & &r(c)&t(a)\} &:0.009   \\
     \{&  s(b, c) &w(b, c) &r(c)&t(a)\} &:0.049   \\
     \{ &  & &r(c)&t(a)\} &:0.021   \\
      \{s(a, b) & s(b,c) & & &t(a)\} &:0.105   \\
     \{s(a, b) &  & & &t(a)\} &:0.045   \\
     \{&  s(b, c) & & &t(a)\} &:0.245   \\
     \{ &  & & &t(a)\} &:0.105   \\
\end{array}\right\}
\end{equation*}
\vspace{0.3cm}

Answering $Q_1$, $Q_2$ leads to the target schema solution $\{v(a, 0.141),$ $v(b, 0.154),$ $u(0.154)\}$. Hence,
the resulting answer set is $\{Q_1(a, 0.141), Q_2(b, 0.154)\}$.
\end{example}

\subsection*{Query Answering in NeuroLang}\label{sec:query-answering}

A {\em NeuroLang query} $Q$
is any conjunction of atoms in $\mathcal{R}_D \cup \mathcal{T}$, such that atoms in $\mathcal{T}$ have as distinguished term a variable; these variables will be instantiated with the probability of certain events as computed by the inference mechanism.
Algorithm~\ref{alg:NeuroLangQA} describes the pseudocode for answering queries in the NeuroLang 
framework---\cref{figure:neurolang} provides a high-level view of the main 
steps involved in this process, where inputs are as defined above.

There are two steps in which {\sc NeuroLangQA} makes external calls.
First, in Step~1 the rewriting of $\Sigma$ w.r.t.\ $\Sigma_1$
is done by means of the XRewrite algorithm developed in~\cite{gottlob_query_2014}
for rewriting queries with respect to the Sticky fragment of existential rules (also known as
Datalog+/--). Note that here the algorithm is used to rewrite
every appearance of heads of rules in $\Sigma_1$ in the bodies
of rules in $\Sigma$, yielding a potentially larger set of
full TGDs (rules without existentials in the head).

Then, Step~3 derives the probabilities associated with atoms.
This is done by dynamically choosing the best algorithm for the job:
if $\pi$ is liftable according to~\cite{dalviDichotomyProbabilisticInference2012}, 
then lifted query answering is applied; otherwise, the query is compiled to an SDD representation and model counting is applied~\citep{VlasselaerPLP14}.
Both cases are implemented in relational algebra with provenance~\citep{senellartProvenanceProbabilitiesRelational2017}.
Note also that up to Step~3 we can guarantee the correctness of the semantics of {\sc NeuroLangQA}, i.e., the probabilities associated with atoms in set $B$ correspond to the probability with which they are entailed in the probabilistic ontology. However, since after this step users can manipulate the probabilities of atoms through aggregation functions provided in $A$, it cannot be guaranteed that this relationship holds in the next steps, so users have the responsibility of making a sound use of such values. This manipulation is intentionally incorporated to increase the expressive power of the languages; similar additions occur in other languages, like Prolog. This feature is useful in our application case allowing, for instance, to aggregate probabilistic values into voxel overlays (cf.\ \cref{sec:forward}), or select the 95th percentile top probabilities of a result set (cf.\ \cref{sec:reverse-cogat}).

The final step of the algorithm returns the answers to query $Q$ as the set of all tuples $t$ built from $\Delta$ such that there exists a homomorphism $\mu$ where $\mu(\vett{X}) = t$ and $\mu(\Phi(\vett{X}, \vett{Y})) \in M'$.

\paragraph{Correctness of {\sc NeuroLangQA}}

We now discuss the correctness of {\sc NeuroLangQA} algorithm with respect to the probabilistic semantics described in Section~\ref{sec:probabilisticmodel}.
Without loss of generality, we assume a query of the form 
\[
Q(\vett{X},\rho_{\vett{X}}) = \Phi(\vett{X}) \wedge \psi_i(\vett{X},\rho_{\vett{X}}),
\]
where
$\Phi(\vett{X})$ is a conjunction of
atoms in $\mathcal{R}_D$
and  $\psi_i(\vett{X},\rho_{\vett{X}})$ is an atom in $\mathcal{T}$.

The result of Step~1 in {\sc NeuroLangQA}
is a special case of a probabilistic ontology $(D', \Sigma')$, where
$\Sigma'$ is a set of full TGDs that may contain stratified negation and recursion.
Furthermore, Step~2a removes from $\Sigma'$
all rules that depend on $C \cup \chi \cup \Phi$~\citep{baget2011rules}.
Therefore, $M$ computed in Step~2b
is unique as neither probabilistic atoms, 
nor existential rules are involved. 
Step~3 now considers the probabilistic ontology defined by $\mathcal{O}= (M, C\cup C', \chi)$.
Note that atoms in $M$ materialize 
ontology $(D',\textit{Aux})$ and they will hold in
every possible world for probabilistic
ontology~$\mathcal{O}$.

Recall that the purpose of PERs is to incorporate the probability of an atom
as an additional term---Step~3 does precisely that: for each PER $\pi$,
it computes the probability of all ground instantiations of $\textit{body}(\pi)$
that are entailed by $\mathcal{O}$.
For each such instantiation $t$, set $B$ contains the instantiation itself ($Q_{\pi}(t)$) and the head of $\pi$
instantiated by values in $t$ and an extra position with value $\textit{Pr}^{\mathcal{O}}(\textit{body}(\pi)(t))$.

Finally, Step~4 considers a deterministic
ontology comprised by $B$ (a set of ground atoms) and the set of 
full TGDs $(\Sigma' \setminus Aux) \cup A$; $M'$ contains all ground atoms that are entailed
by such ontology. As in the case of $M$, $M'$ is unique
since neither existential rules nor probabilistic atoms are involved.
 
Therefore, we can conclude that---by construction---the results computed by the 
{\sc NeuroLangQA} algorithm are correct with respect to the probabilistic semantics 
defined in Section~\ref{sec:probabilisticmodel} up to Step~3. This means that the 
probabilities associated with atoms in $B$ correspond to the probability with which 
they are entailed by the probabilistic ontology.
The final two steps simply follow the user-specified rules for establishing personalized 
views, which may manipulate probability values in an arbitrary fashion.
With the framework in place, in the following we show how it can be 
applied in practice.

\section{Evaluation based on Real-World Use Cases in Neuroscience Research}
\label{sec:use-cases}
In this section, we illustrate via concrete examples several use cases that appear in real-world tasks carried out by neuroscience researchers. Since all of our analyses are based on meta-analytic components, we first give a brief description of the Neurosynth database we use in our examples. Where extra data is used, it will be clarified in each particular case.
The Neurosynth database is composed of $\num{3228}$ terms, $\num{14370}$ studies (\textit{SelectedStudy}), and $\num{33593}$ voxels; but this information would not be useful without associations, so we also have $\num{1049299}$ terms reported as present in studies (\textit{TermInStudy}) and $\num{507891}$ voxels reported as active (\textit{FocusReported}), also with their respective study. Finally, there are 112 brain regions from Destrieux's atlas~\citep{destrieux_automatic_2010} associated with brain coordinates through the \textit{VoxelByRegionDestrieux} relation. These data give rise to the following extensional databases:

\begin{equation*}
    \begin{aligned}
        D&=\left\{ \begin{array}{l}
        \textit{TermInStudy}(\textit{``emotion"}, s_1),\\ 
        \smash{\vdots} \\
        \textit{TermInStudy}(\textit{``pain"}, s_{120}),\\
        \textit{FocusReported}(5, -5, 3, s_1),\\
        \smash{\vdots} \\
        \textit{FocusReported}(-10, 5, 1, s_{25}),\\
        \textit{VoxelByRegionDestrieux}(15, 47, 16,\\
        \hspace{10pt}\textit{'l\_g\_and\_s\_frontomargin'}),\\
        \smash{\vdots} \\
        \textit{VoxelByRegionDestrieux}(16, 46, 15,\\
        \hspace{10pt}\textit{'l\_g\_and\_s\_frontomargin'}),\\
        \end{array}\right\}\\[6pt]
        C &= \left\{ \begin{array}{l}
        \textit{SelectedStudy}(s_i):\frac{1}{\# studies} \\
        \textit{FocusCoactivates}(\,5, -5, \,3,\quad \, 5, -5, \,3):1 \\
        \smash{\vdots} \\
        \textit{FocusCoactivates}(\,5, -5, \,3,\quad -10, \,\,\,5, \,1): \\
        \quad\quad (2\pi 2)^{-3/2}\exp\left(-\frac 1 2\frac{\|(5, -5, 3)-(-10, 5, 1)\|^2}{2^2}\right)\\
        \end{array}\right\}\\
    \end{aligned}
    \vspace{0.3cm}
\end{equation*}
where $\textit{FocusCoactivates}$ represents spatial uncertainty in foci reporting, as they encode that the probability that two foci co-activate is mediated by their distance as measured by a 3D Gaussian law with standard deviation 2mm. This dataset has approximately~5 million atoms. Furthermore, the CogAt ontology~\citep{poldrackCognitiveAtlasKnowledge2011} is composed of $\num{56807}$ rules. In the following, 
examples are written in extended Datalog syntax, as in our implemented tool\footnotemark. 
We base our examples on versions 1.4.0 of IOBC, 0.3.1 of CogAt, and the Destrieux 2009 atlas~\citep{destrieux_sulcal_2009} provided by Nilearn software package v0.7.0~\citep{fischl_automatically_2004}. In addition, both the software code and other examples can be found on the official NeuroLang repository\footnotemark[\value{footnote}].

\footnotetext{https://neurolang.github.io/}
% Neurosynth database:
% Terms: 3228
% Voxels: 33593
% Studies: 14370 (SelectedStudy)

% Voxels reported in studies: 507891 (FocusReported)
% Terms reported in studies: 1049299 (TermInStudy)

\subsection{Open world assumption}\label{sec:open-world}

We now show how we can make use of NeuroLang to solve queries that require taking into account the open world assumption. For that purpose, we use the terms present in the Neurosynth database to associate the studies analyzed with the cognitive processes proposed by CogAt. For this, we make use of a special term included by our ontologies parser, \textit{Entity(t, s)}, that will allow us to associate external data with the internal entities of the ontology. 

\begin{listing}
\begin{lstlisting}[caption={Example of rules introduced by our parser when processing ontologies, allowing us to associate internal entities with external datasets.},label={lst:open_world_entity}]
SpatialAttention(X) :- Entity('spatial attention', x)
 \end{lstlisting}
\end{listing}

Each entity that is parsed creates this specific rule, that we can later overload with external information, creating an association between entities and external data. An example of this overloading process can be seen in the first line of \cref{lst:query_visualawareness} where we associate the studies of the Neurosynth database with the entities of the ontology, allowing us to perform queries that return these studies, but under the universe modeled by the ontology. All within the same semantics of the NeuroLang program.

As we can see in the first line of \cref{lst:query_visualawareness}, we then associate these entities with the Neurosynth studies within the same program. This will allow us to combine both datasets, so that we can use the information structured in the CogAt ontology to ask questions that result in Neurosynth's studies. 

\begin{listing}
\begin{lstlisting}[caption={Open world assumption. See the description in \cref{sec:open-world}}, label={lst:query_visualawareness}]
Entity(t, s) :- TermInStudy(t, s)

OpenWorldStudies(s) :- PartOf(t, s), VisualAwareness(s).

ProbMap(x, y, z, PROB) :- FocusReported(x, y, z, s) // 
    (SelectedStudy(s) & OpenWorldStudies(s))
    
ProbabilityImage(create_region_overlay(x, y, z, p)) :- 
    ProbMap(x, y, z, p)
\end{lstlisting}
\end{listing}

We focus on solving queries based on some of the ontology's constraints defining open-world knowledge. In particular, we aim at relating the \textit{visual awareness} cognitive process from the CogAt ontology with brain areas reported activate during this process. This can't be done directly through Neurosynth, as the cognitive process is not reported. Therefore, we need a way to associate studies related to this term that don't mention \textit{visual awareness} explicitly. CogAt helps in solving this problem: there is TGD specifying that \textit{spatial attention} is a sub-process of \textit{visual awareness}. Which, expressed as a Datalog+/- rule in CogAt’s TGD set, is: \begin{equation}
  \begin{aligned} \label{eq:owl_fol}
   &\forall X \text{SpatialAttention}(X)\rightarrow \\ 
   & \quad\quad\quad \exists Y\text{PartOf}(X, Y)\wedge \text{VisualAwareness}(Y)
  \end{aligned} \in \Sigma_1^{\text{CogAt}},
\end{equation} which has an existentially-quantified variable in the head, hence representing open-world knowledge.

\begin{figure}[t]
    \centering
    \includegraphics[scale=0.5]{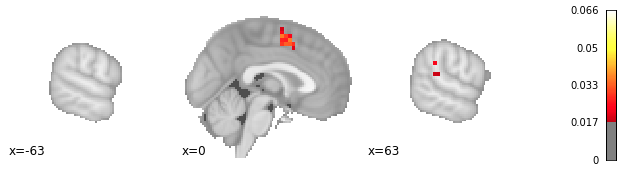}
    \caption{Resulting thresholded brain image from the NeuroLang use case showing the activations related to spatial attention obtained through the resolution of a query under the open world assumption.}
    \label{ex:open-world}
\end{figure}

We seamlessly harness this open knowledge to analyse activations related to \textit{visual awareness} using to NeuroLang's built-in capabilities: we write a program (see \cref{lst:query_visualawareness}) to obtain all studies that, while not mentioning \textit{visual awareness}, mention terms which, according to CogAt, imply that the cognitive process is involved. Importantly, we achieve this by combining an automatically-produced literature database with a expert-produced ontology. The resulting activations can be seen in \cref{ex:open-world}

\subsection{Forward inference}\label{sec:forward}

In this task, we wish to assess the probability of a voxel being
reported as active in a study given that the word ``emotion''
is present in the specific study. 

\begin{listing}
\begin{lstlisting}[caption={Forward inference}]
TermAssociation(t) :- SelectedStudy(s), 
    TermInStudy(t, s).

Activation(i, j, k) :- SelectedStudy(s), 
    FocusReported(i1, j1, k1, s),
    FocusCoactivates(i, j, k, i1, j1, k1).

% Probability Encoding Rule: PROB is 
% used to encode probability as defined in 
% Section 3. The // operator is 
% syntactic sugar for conditional 
% probability as P(A|B) = P(A,B) / P(B).
ProbMap(i, j, k, PROB) :-
    Activation(i, j, k)
    // TermAssociation("emotion").

% Aggregation to build a single image with
% the probability p in each position
% within the top 95% of probability
Percentile_95(compute_percentile(p, 95)) :-
    ProbMap(i, j, k, p).
    
ProbabilityImage(create_region_overlay(i, j, k, p)) :-
    ProbMap(i, j, k, p), 
    Percentile_95(p95), p > p95

Ans(x) :- ProbabilityImage(x)
\end{lstlisting}
\end{listing}

\begin{figure}[H]
    \centering
    \includegraphics[scale=0.5]{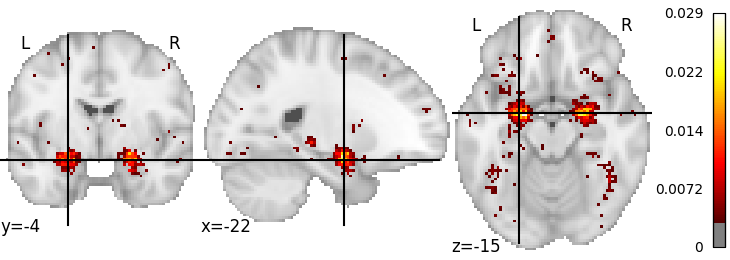}
    \caption{Resulting thresholded brain image from the NeuroLang use case showing that foci in the amygdala are most probably reported if a study includes the word ``emotion''. As expected, the main area shown corresponds to the amygdala~\citep{mesulamSensationCognition1998}.}
    \label{ex:amygdala}
\end{figure}

Note that in order to represent this knowledge we only need the expressive power of full TGDs (no existential rules are needed).
In \cref{ex:amygdala} we see that the most important reported activations are concentrated in the amygdala, the region most related to emotions, as generally accepted in the neuroscience~field.

\subsection{Reverse inference over a region of the Destrieux atlas leveraging the CogAt ontology}\label{sec:reverse-cogat}

For this use case, we will use reverse inference techniques to obtain the terms most likely to be associated with the \textit{short insular gyrus} of the Destrieux atlas. {\em Atlases} are parcellations of the brain into distinct areas based on histological, physiological, or other characteristics. In addition, we will also use the information stored in the CogAt ontology to filter the terms from the reverse inference in order to obtain cleaner results, all in the same query. Terms included in the CogAt ontology are characterized by the ``label'' relation, which we load into our system under the CogAtLabel symbol. The CogAt ontology rewriting adds $\num{4577}$ formulas to our database. The code of this program is presented in \cref{lst:reverse}.

\begin{listing}
\begin{lstlisting}[caption={Reverse inference over a region of the Destrieux atlas leveraging the CogAt ontology.}, label={lst:reverse}]
FilteredTerms(s, t) :- TermInStudy(s, t), CogAtLabel(uri, t).

RegionActivated(s) :- 
    VoxelByRegionDestrieux(i, j, k, "l_g_insular_short"), 
    FocusReported(i1, j1, k1, s), 
    FocusCoactivates(i, j, k, i1, j1, k1).

RegionActivated(s) :- 
    VoxelByRegionDestrieux(i, j, k, "r_g_insular_short"),
    FocusReported(i1, j1, k1, s),
    FocusCoactivates(i, j, k, i1, j1, k1).

TermProbability(t, "unfiltered", PROB) :- TermInStudy(s, t) // 
    (RegionActivated(s), SelectedStudy(s)).
    
TermProbability(t, "filtered", PROB) :- FilteredTerms(s, t) // 
    (RegionActivated(s), SelectedStudy(s)).

Percentile_95(is_filtered, compute_percentile(p, 95)) :- 
    TermProbability(t, is_filtered, p).
    
Ans(t, is_filtered, p) :- TermProbability(t, is_filtered, p), 
    percentile_95(is_filtered, p95), p > p95.
    
\end{lstlisting}
\end{listing}

We can see in \cref{tab:destrieux-non-filtered} (right) how, by using the knowledge stored in the CogAt ontology, we can filter out those terms that, being present in most neuroimaging studies, only add noise to the results. Therefore, we obtain a list of much more relevant results that are also more closely related to the general knowledge of the field of neuroscience. Solving this query takes approximately~6 seconds. For another use case leveraging ontological knowledge, please refer to \cref{sec:ontology-synonyms}.

\begin{table}
    \centering
    \begin{minipage}[t][][t]{.49\linewidth}
    \centering
        \begin{tabular}{lr}
        \multicolumn{2}{c}{{\bf Unfiltered results}}\\ 
        \toprule
                         {\em Term} & {\em Prob} \\
        \midrule
                         task &  0.47 \\
                     magnetic &  0.47 \\
                    resonance &  0.47 \\
           magnetic resonance &  0.47 \\
          functional magnetic &  0.43 \\
                        using &  0.38 \\
                      frontal &  0.37 \\
                     anterior &  0.35 \\
                      network &  0.34 \\
                   prefrontal &  0.33 \\
        \bottomrule
        \end{tabular}
    \end{minipage}
    \begin{minipage}[t][][t]{.5\linewidth}
        \begin{tabular}{lr}
        \multicolumn{2}{c}{{\bf Filtered results}}\\ 
        \toprule
                   {\em Term} & {\em Prob} \\
        \midrule
                 memory &  0.20 \\
              attention &  0.14 \\
         working memory &  0.09 \\
             perception &  0.09 \\
               learning &  0.08 \\
               language &  0.08 \\
                emotion &  0.07 \\
                        &       \\
                        &       \\
                        &       \\
        \bottomrule
        \end{tabular}
    \end{minipage}
    \vspace{0.3cm}
    \caption{Results from \cref{sec:reverse-cogat}. {\em Left}: Results (10 of 161 most relevant terms in the top 0.5\% most probable terms) of applying reverse inference on region \textit{short insular gyri} of the Destrieux atlas using Neurosynth term association. Results include irrelevant results in terms of cognitive tasks such as ``magnetic resonance''. 
    {\em Right}: Same approach, but filtering of terms based on those present in the CogAt ontology~\citep{nieuwenhuysInsularCortexReview2012}.}
    \label{tab:destrieux-non-filtered}
\end{table}

\subsubsection{Retrieving information from related terms via the hierarchical structure of the ontology}\label{sec:ontology-synonyms}

We now show how we can leverage
the ontological knowledge provided by the International Organization for Biological Control (IOBC) to perform an analysis that includes terms related to our main term (\textit{noxious} and \textit{nociceptive} related to \textit{pain}, in this example) without knowing them beforehand, enriching our analysis automatically. The IOBC ontology rewriting adds~11,102 formulas to our database.

\begin{listing}
\begin{lstlisting}[caption={Retrieving information from related terms via the hierarchical structure of the ontology}, label={lst:iobc_pain}]
RelatedTerm(term) :- term == "pain".

RelatedTerm(term) :- label(pain_entity, "pain"), 
    altLabel(subclass, term),
    related(pain_entity, subclass).

FilteredBySynonym(t, s) :- TermInStudy(t, s), RelatedTerm(t).

Result(i, j, k, PROB) :- FocusReported(i, j, k, s) // 
    (SelectedStudy(s), FilteredBySynonym(t, s)).

Percentile_95(compute_percentile(p, 95)) :- Result(i, j, k, p).

VoxelActivationImg(create_region_overlay(i, j, k, p)) :-
    Result(i, j, k, p),
    Percentile_95(p95), p > p95.

ans(img) :- VoxelActivationImg(img).
\end{lstlisting}
\end{listing}

\begin{figure}[H]
    \centering
    \includegraphics[scale=0.5]{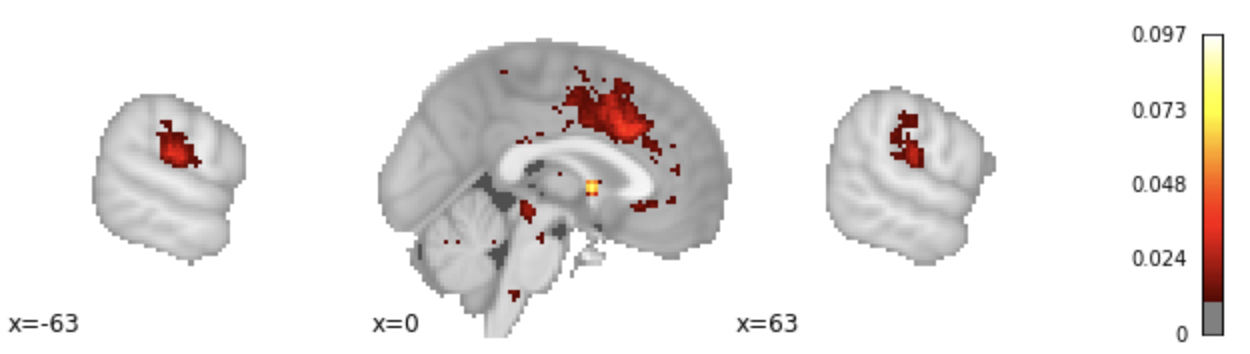}
    \caption{Resulting thresholded brain image from the NeuroLang use case showing the activations related to pain and its related terms derived from the IOBC ontology (noxious and nociceptive). Dorsal anterior cingulate cortex (x=0) and parietal regions are be active in articles mentioning pain and related words. agreeing with current knowledge in pain location~\citep{liebermanDorsalAnteriorCingulate2015}.}
    \label{ex:synonyms}
\end{figure}

In~\cref{ex:synonyms}, we provide a view of the results obtained from this example (see \cref{lst:iobc_pain}). In this case, the activations of Noxious and Nociceptive were also automatically included in the result, solving one of the current problems of Neurosynth (the need to know all the terms you want to use beforehand). Solving this query takes approximately~55 seconds.

\subsection{Segregation reverse inference query}\label{sec:segregation}

This final use case shows how we can use negation and existentials to express specificity.
We pick the terms present in the CogAt ontology that are mentioned in studies reporting activations within the short insural gyri.

\begin{listing}
\begin{lstlisting}[caption={Segregation reverse inference query. See the description in \cref{sec:segregation}.}, label={lst:segregation}]
OntologyTerms(t) :- hasTopConcept(u, c), label(u, t)
FilteredTerms(s, t) :- TermInStudy(s, t), OntologyTerms(t)
RegionActivated(s, r) :- VoxelByRegionDestrieux(i, j, k, r), 
    FocusReported(i, j, k, s).
SegregatedStudies(s) :- RegionActivated(s, r), 
    (DestrieuxLabels(r, 'l_g_insular_short') | 
    DestrieuxLabels(r, 'l_g_insular_short')),
    ~exists(r2, RegionActivated(s, r2), r != r2)
TermProbability(t, PROB) :- FilteredTerm(s, t)
    // (SegregatedStudies(s)
      SelectedStudy(s)).
Percentile_95(compute_percentile(p, 95)) :- TermProbability(t, p).
Ans(t, p) :- TermProbability(t, p), Percentile_95(p95), p > p95.
\end{lstlisting}
\end{listing}
\vspace{0.3cm}

Processing took 42.45 seconds. Results are shown in \cref{table:segregation}.

\begin{table}[h]
    \centering
    \begin{tabular}{lr}
    \toprule
              term &      prob \\
    \midrule
           anxiety &  0.097819 \\
    \bottomrule
    \end{tabular}
    \vspace{0.2cm}
    \caption{Terms, within the 95th percentile, mentioned in our segregation query in \cref{sec:segregation}. Shows that studies presenting activations only related to the short insula gyrus tend to be associated with anxiety.} \label{table:segregation}
\end{table}

\section{Discussion and Conclusion}
\label{sec:Disc-concl}
In this paper, we presented a fragment of probabilistic Datalog+/-- enriched with negation and aggregation, along with a scalable query resolution algorithm. The main goal of our specific approach is meta-analysis of neuroimaging data. Several different approaches to probabilistic Datalog+/-- semantics and query resolution   exist~\citep{gottlob2013query,ceylanOpenworldProbabilisticDatabases2021}. Nonetheless, these do not incorporate aggregation, and the possibility of manipulating the probabilistic query results within the same language. These two features, as shown by our use-case analysis in Section~\ref{sec:use-cases}, are fundamental traits required to provide a probabilistic logic programming language that can encode neuroimaging meta-analysis applications end-to-end.

The possibility of manipulating probabilities within the language comes at a great expense. After our PERs are computed, in Step~4 of Algorithm~\ref{alg:NeuroLangQA}, our language allows handling probabilities as a standard float column. While this allows for analyses required by our target applications, it calls for disciplined programming from the user such that the manipulation of probabilities remains sound. Nonetheless, this gives our language great power; for instance, we can build probabilistic brain images, through aggregation, as shown in Sections~\ref{sec:forward}--\ref{sec:reverse-cogat}; and compute the probability differences between two events, which we show in Section~\ref{sec:segregation}.

All these features allow us to go beyond current tools in meta analyses whose queries are based in propositional logic~\citep{yarkoniLargescaleAutomatedSynthesis2011,lairdBrainMapStrategyStandardization2011} and harness the full power of the FO$^{\neg\exists}$ fragment, as well as open-world semantics, to express meta-analysis tasks in a sound, disciplined, and declarative manner. Furthermore, by using, as in~\cite{ceylanOpenworldProbabilisticDatabases2021}, a lifted query processing approach when possible (see Algorithm~\ref{alg:NeuroLangQA}, Step~3), we are able to process current meta-analytic datasets enriched with ontologies that are of considerable size, as described at the beginning of Section~\ref{sec:use-cases}. While it is true that there are other works that make possible the resolution of Datalog+/-- queries~(\cite{ceylanOpenworldProbabilisticDatabases2021, jha_probabilistic_2012}), the definition of the problem we wish to solve makes it necessary to have a framework capable of solving probabilistic choice and handling deterministic open-world knowledge. Moreover, we are not aware of any practical implementation of the mentioned works, beyond the provided theory.

To conclude, we have shown that neuroimaging meta-analytic applications are an excellent real-world application for a language such as probabilistic Datalog+/--. By using probabilistic semantics that have recently converged from different probabilistic logic and open-world language approaches~\citep{riguzziALLPADApproximateLearning2008,ceylanOpenworldProbabilisticDatabases2021,vennekensCPlogicLanguageCausal2009}, with open-world semantics~\citep{CaliGL12,gottlob_query_2014,ceylanOpenworldProbabilisticDatabases2021}, and query resolution approaches~\citep{dalviDichotomyProbabilisticInference2012,ceylanOpenworldProbabilisticDatabases2021,vlasselaerKnowledgeCompilationWeighted}, we have produced a language that is ready to be used in neuroimaging applications.

\section*{Statements and Declarations}
The authors have no relevant financial or non-financial interests to disclose. 
The authors declare that they have no competing interest.

\section*{Information Sharing Statement}
All the datasets and software used in this article are openly available at the following web sites:
\begin{itemize}
    \item Neurolang: https://neurolang.github.io/
    \item Nilearn, version 0.7.0: https://nilearn.github.io/
    \item NeuroSynth database: https://github.com/neurosynth/neurosynth
    \item CogAt ontology, version  0.3.1: \\
        https://bioportal.bioontology.org/ontologies/COGAT
    \item IOBC ontology, version  1.4.0: \\
        https://bioportal.bioontology.org/ontologies/IOBC
\end{itemize}

\section*{Acknowledgement}
This work was partially supported by the ERC-StG NeuroLang ID:757672. We are deeply thankful to the NiLearn community for the data ingestion and visualization tools~\citep{abrahamMachineLearningNeuroimaging2014}.
\clearpage

\bibliography{biblio, biblio_demian, references}

\end{document}